\documentclass[conference]{IEEEtran}

\usepackage{tabularx}
\usepackage{algorithm}
\usepackage{algorithmic}
\usepackage{enumitem}
\usepackage{cleveref}

\usepackage{graphicx}
\usepackage{amsmath}
\ifCLASSINFOpdf
\else
\fi
\begin{document}
%
\title{DeepCancer: Detecting Cancer through Gene Expressions via Deep Generative Learning}

\author{\IEEEauthorblockN{Rajendra Rana Bhat\textsuperscript{*}, Vivek Viswanath\textsuperscript{*}, Xiaolin Li\textsuperscript}
\IEEEauthorblockA{Large-scale Intelligent Systems Laboratory\\
Department of Electrical and Computer Engineering\\
University of Florida, Gainesville, Florida 32603--0250\\
\{rbhat,vivek201189\}@ufl.edu, andyli@ece.ufl.edu}
\textit{* Equal contribution.}
}


%


\maketitle

\begin{abstract}
Transcriptional profiling on microarrays to obtain gene expressions has been used to facilitate cancer diagnosis. We propose a deep generative machine learning architecture (called DeepCancer) that learn features from unlabeled microarray data. These models have been used in conjunction with conventional classifiers that perform classification of the tissue samples as either being cancerous or non-cancerous. The proposed model has been tested on two different clinical datasets. The evaluation demonstrates that DeepCancer model achieves a very high precision score, while significantly controlling the false positive and false negative scores.
\end{abstract}

\vspace{2mm}


%
\IEEEpeerreviewmaketitle

\section{Introduction}
Detection of cancer has always been at the forefront of all research endeavors within the medical community. Advancements in medical sciences and technology have given birth to numerous methodologies that leverage the availability of massive amounts of data to provide valuable insight into several cancer-related processes~\cite{Kourou}, despite the complexity associated with these methodologies. Gene expression profiling technologies comprise one such research avenue that relies on establishing correlations between different gene-expression data, which are generated through oligonucleotide arrays or cDNA microarrays, and the different cell stages observed in cancer patients, thus enriching cancer taxonomy~\cite{Amir}~\cite{Therese}. Owing to this ability, this technology has been widely applied to study and enhance diagnosis of breast cancer and prostate cancer. Inflammatory breast cancer literature~\cite{Dawood} reports that 1-5\% of all cancer-based mortalities in the United States in 2011 were due to the highly rare inflammatory breast cancer which is known to be exhaustively difficult to diagnose, while prostate cancer has been predicted to occur in about one-quarter of all male cancer diagnoses\cite{Miller}, thereby impressing an urgent need for more reliable diagnosis techniques. Current literature is rife with instances of thorough research of gene signatures in attempting to distinguish genes expressed in affected (cancerous) tissues from those expressed in normal (non-cancerous) tissues, yet high data dimensionality is an inherent characteristic of these microarray experiments, thereby persistently posing a serious challenge to reliable and effective diagnosis. \\
\indent There are several machine learning models that address this challenge. Since these models learn from experience, reliable classification demands extensive training on data that are usually heterogeneous in that real-world data samples belong to disparate classes with a high likelihood of redundancy in data, thus resulting in high dimensionality. Therefore, these models have been designed and/or improved to tackle the curse of dimensionality~\cite{Larochelle} in a way that they are suitable for scenarios involving high dimensions. Deep machine learning \cite{Yoshua}, a specialized subset of machine learning, presents a class of data-friendly models that are highly efficient in learning hidden features intrinsic to the data based on a multilayered architecture, with lower layers learning simpler features and eventually composing them into more complex ones in subsequent layers in either a supervised or an unsupervised fashion.\\
\indent DeepCancer explores one such model called Generative Adversarial Networks~\cite{Goodfellow} that involve a generative network and an inference network functioning in an adversarial learning setup. The generative network learns to probabilistically generate output samples, given random noise as input, whereas the inference or discriminator network learns to discriminate the ‘true’ data distribution samples from the generated ‘fake’ data distribution samples. This way, the generator tries to fool the discriminator by trying to generate progressively better ‘fake’ samples, and the discriminator tries best not to be fooled by the generator by improving its ability to classify samples as either being ‘real’ or ‘fake’, until the generated samples are indistinguishable from the original samples. These features learnt by the discriminator from gene expression microarray data can finally be passed through sigmoid activation for supervised classification of gene samples as either being cancerous or non-cancerous. We test the ability of our model in classifying breast cancer and prostate cancer samples and show that this hybrid model performs this classification task accurately. Also, as a baseline, we implement a standard Restricted Boltzmann Machine (RBM)~\cite{Rumelhart} that works, similar to these adversarial networks, in a pipeline with a traditional classifier.

\section{Related Work}

Gene expressions have classically been researched quite extensively through employing machine learning (ML) techniques in the context of both supervised and unsupervised methods of training. Unsupervised clustering, including its variants such as hierarchical or k-means clustering~\cite{Wen}~\cite{Weinstein-John}~\cite{Michaels}~\cite{Sokal}, have been widely used methods to analyze gene expression data~\cite{Javier}~\cite{Asir}~\cite{Crampin}~\cite{Nilamani}~\cite{Chun}~\cite{Berrar}. Clustering based on metrics like correlation coefficients was explored in~\cite{Eisen} in order to represent relationships among genes in a tree structure with the branch length signifying the degree of mutual similarity computed by a pairwise average-linkage cluster analysis similarity function. Affinity propagation clustering algorithm, apart from spectral clustering, has been explored in~\cite{Vasilis} wherein a pairwise similarity matrix is the input to the algorithm when the number of clusters is unknown. Multivariate Gaussian mixture models have also been used as the fundamental principle to perform unsupervised clustering of breast cancer samples~\cite{Chris}, to perform supervised clustering analysis of gene expressions of yeast cell data~\cite{Shizhong}. Additionally, Euclidean distance has also been utilized as a metric for performing gene clustering~\cite{Wen}.\\
\indent On the other hand, historically, artificial neural networks (ANNs) have been employed in a supervised framework to analyze gene-expression signatures of small, round blue cell tumors generated by cDNA microarrays and perform diagnostic classification~\cite{Javed}. ANNs have also been applied in esophageal cancer research in order to analyze cDNA microarray data~\cite{Yan}, to distinguish renal cell carcinomas from normal renal cysts~\cite{Philip}, and also in the prediction of colorectal cancer~\cite{Leonardo}. Another variant of ANNs called self-organizing maps (SOMs)~\cite{Kohonen}~\cite{Teuvo} have been equally exploited for carrying out clustering of gene signatures~\cite{Gavin}~\cite{Chun}, although in an unsupervised manner. At the heart of this algorithm is the concept of centroid-based clustering~\cite{Todd}, i.e., the algorithm creates as many centroids as defined by the user by using the data points yielding an optimal set of centroids, which leads to multiple clusters consisting of these `nearby' data points. Support vector machines (SVMs) have been employed because of their ability to choose sparse solutions in high multidimensional feature spaces while rejecting outliers in functionally classifying gene expression data from DNA microarray hybridization experiments~\cite{Kourou}~\cite{Michael}~\cite{Mukherjee}. The fundamental notion involves a margin maximization that effectively separates samples belonging to different classes while respecting generalization to perform highly reliable classification of new (unseen) samples.\\
\indent In this paper, we present an adversarial model that generatively learns features from the true data distribution, thereby improving the discriminator's classification efficiency. This discriminator's features are passed through sigmoid activations prior to classifying a sample as either being cancerous or `normal'. Also, to realize the efficacy of this adversarial network, we propose two models as baseline: an RBM that is trained in a pipeline with a traditional Logistic classifier and another with the SVM.

\section{Generative Learning Models}

\subsection{Restricted Boltzmann Machines}

\begin{figure}[!ht]  
\centering
\includegraphics[scale=0.65]{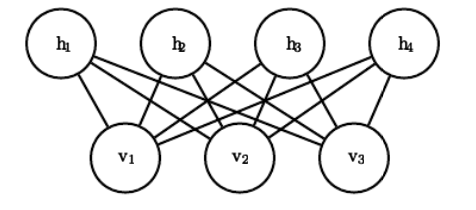}  
\caption{A Restricted Boltzmann Machine Markov network~\cite{Bengio_book}}
\label{fig:rbm1}
\end{figure}

Restricted Boltzmann Machines (RBMs) are structured probabilistic, energy-based, generative, undirected graphical models that learn probability distributions over the set of the provided inputs. As shown in Fig. 1, these networks typically consist of binary multiple visible layers and a single latent or `hidden' layer, and are characterized by a bipartite graph design, i.e., direct interactions between any variables in the visible layer or between hidden units is restricted, thus justifying its nomenclature~\cite{Geoffrey}~\cite{Bengio_book}. These layers can be stacked one on top of the other for designing deeper models. This model involves determining the `energy' or the joint probability distribution of a pair of a visible and hidden unit (v and h respectively) as per the energy function~\cite{Bengio_book} given by:

\begin{equation} \label{eq:equation1}
P(v,h) = \frac{1}{Z}exp\{-E(v,h)\}
\end{equation}

\noindent where $b, c,$ and $W$ are unconstrained, real-valued, learnable parameters and $Z$ is the normalizing constant known as the partition function:

\begin{equation} \label{eq:equation2}
Z = \sum_v \sum_h  exp\{-E(v,h)\}
\end{equation}

\noindent Also, the energy function of an RBM, $E(v,h)$, parameterizes the relationship between the visible and hidden variables:

\begin{equation} \label{eq:equation3}
E(v,h) = -b^Tv -c^Th - v^TWh
\end{equation}

\noindent This translates directly to the following free energy formula:

\begin{equation} \label{eq:equation4}
F(v) = -b'v - \sum_i log\sum_{h_i} e^{h_i(c_i+W_iv)}
\end{equation}

\noindent Furthermore, the gradient of the log probability of a training vector with respect to these weights comes from Eq. \ref{eq:equation1}

\begin{equation} \label{eq:equation5}
\frac{\partial logP(v)}{\partial w_{ij}} = \langle v_i h_j \rangle_{data} - \langle v_i h_j \rangle_{model} 
\end{equation}

\noindent where the angle brackets denote expectation under the distribution specified by the accompanying subscript. Since, there are no direct connections between the hidden units and within the visible units, we get unbiased samples of $\langle v_i h_j \rangle_{data}$, for hidden and visible units respectively, according to:

\begin{equation} \label{eq:equation6}
P(h_i=1|v) = \sigma(\sum_j w_{ij}v_i+c_j )
\end{equation}

\begin{equation} \label{eq:equation7}
P(v_i=1|h) = \sigma(\sum_j w_{ij}h_j+b_i )
\end{equation}

\noindent The two terms in Eq. \ref{eq:equation5} are referred to as the positive and the negative phase, where the terms 'positive' and `negative' refer not to the signs but their effect on the probability density of the model. The first term increases the probability of the training data (by reducing the free energy), whereas the second term has the opposite effect. However, the term $\langle v_i h_j \rangle_{model}$ is more intractable. Therefore, a standard practice is to numerically approximate by drawing samples from the data distribution. Then, a reconstruction is done in three simple steps:
\begin{enumerate}
\item Assign training vector to visible units
\item Compute hidden unit states using Eq. \ref{eq:equation6}
\item Compute reconstruction by setting each $v_i$ to 1 with probability given in Eq. \ref{eq:equation7}.
\end{enumerate}

\noindent The resultant change in weights is similar to Eq. \ref{eq:equation5}:

\begin{equation} \label{eq:equation8}
\Delta w_{ij} = \epsilon (\langle v_i h_j \rangle_{data} - \langle v_i h_j \rangle_{recon})
\end{equation}

\noindent where $\epsilon$ is the learning rate.  This technique is called Contrastive Divergence and its approximation is a difference mathematically written as:

\begin{equation} \label{eq:equation9}
CD_n = KL(p_0||p_{\infty}) - KL(p_n||p_{\infty})
\end{equation}

\noindent where, KL(.) is the Kullback-Liebler divergence for the model $p_{\infty}$ and for the data $p_0$.

\subsection{Generative Adversarial Networks}
A fairly recent and far less researched model, generative adversarial networks~\cite{Goodfellow} (GANs) in the context of deep machine learning are generative models involving two networks competing against each other -- a generator that tries to mimic examples from the training dataset, which is sampled from the true data distribution. The discriminator receives samples from the without being told where the sample came from. Therefore, it tries to predict whether a sample is `true' or synthetic. This way, the discriminator tries to classify samples as accurately as possible, while the generator tries to fool the discriminator into thinking the sample came from the true data. This two-player minmax game, thereby, improves the ability of both the networks to perform their task as accurately as they can, until the `fake' samples are virtually indistinguishable from the original samples.

Mathematically, as explained in~\cite{Goodfellow}, a prior on input noise variables $p_z(z)$ is defined to learn the generator's distribution $p_g$ over data $x$. This prior is then used to represent a mapping to the data space as $G(z;\theta_g)$  , where $G$ is a differentiable function represented by a multilayer perceptron. Similarly, $D(x,\theta_d)$  represents the probability of $x$ belonging to the true data distribution. Therefore, the minmax game of the model defines a loss function:
\begin{equation} \label{eq:equation10}
\begin{split}
\underset{G}{\text{min}} \ \underset{D}{\text{max}}V(D,G)= E_{x\sim p_{data}(x)}[logD(x)]								\\+ E_{z\sim p_z(z)}[log(1-D(G(z)))]
\end{split}
\end{equation}

\noindent where, $D$ is maximized to assign the correct label to both the true and the generated samples, and $G$ is trained to minimize the logarithm term, $log(1-D(G(z)))$ . This is achieved through k stochastic gradient ascent update steps on the discriminator as:

\begin{equation} \label{eq:equation11}
\bigtriangledown_{\theta_d}\frac{1}{m}\sum_{i=1}^m[logD(x^{(i)})+log(1-D(G(z^{(i)})))]
\end{equation}

\noindent The generator updates through only one step of stochastic gradient descent:

\begin{equation} \label{eq:equation12}
\bigtriangledown_{\theta_g}\frac{1}{m}\sum_{i=1}^m log(1-D(G(z^{(i)})))
\end{equation}

\noindent Ideally, after several updates epochs, for a fixed G, at optimality,

\begin{equation} \label{eq:equation13}
 D_G^*(x) = \frac{p_{data}(x)}{p_{data}(x)+p_g(x)}
\end{equation}

\noindent where $D^*$ denotes the optimal discriminator.

\begin{figure*}[!htb]  
\centering
\includegraphics[scale=0.50]{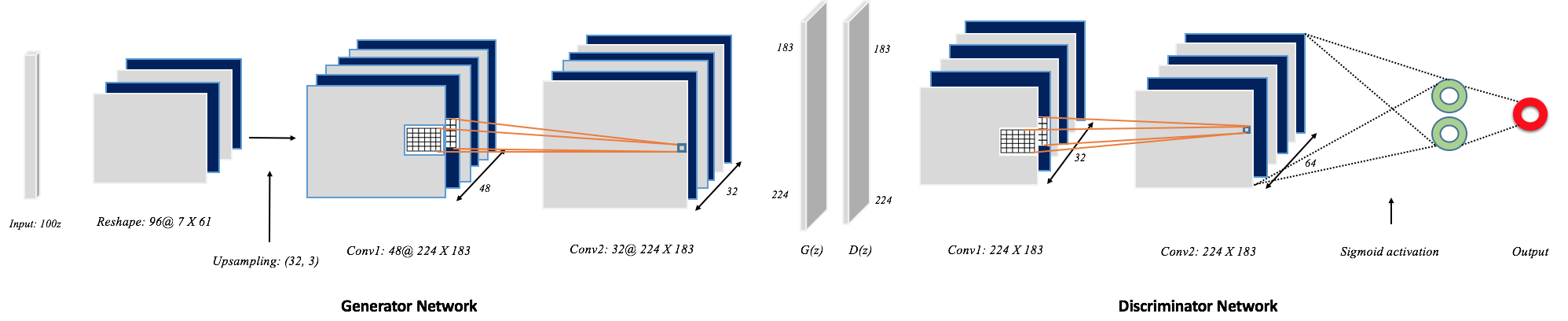}  
\caption{Fig. 2 Deep Convolutional architecture for the adversarial network. The dimensions shown are for the GEO ID GSE45584 dataset. The corresponding parameters for the prostate cancer task are mentioned in Table~\ref{table:ganarch}}
\label{fig:gan}
\end{figure*}



\begin{algorithm*}[!htb] 
\label{alg:alg1}
\caption{Deep Convolutional Model for adversarial training. The generator--discriminator pair learns features through convolutions over input matrix. $k$ is a hyperparameter for the discriminator. $k=1$ proved to be sufficient.}
\begin{algorithmic}
\FOR{$\mbox{number of epochs}$}
   \FOR{$\mbox{$k$ steps}$} 
   \STATE{\mbox{Sample minibatch of noise examples from the standard uniform distribution}} 
   \STATE{\mbox{Sample minibatch of examples from the data distribution}}
   \STATE{\mbox{Perform convolutions on the input}}
   \STATE{\mbox{Flatten the convolution ouput and pass through sigmoid}}
   \STATE{\mbox{Compute gradients through the optimization function}}
   \STATE{\mbox{Freeze the discriminator traning}}
   \ENDFOR
   
   \STATE{\mbox{Sample minibatch of noise examples from the standard uniform distribution}}
   \STATE{\mbox{Reshape and upsample the minibatch for convolutions}}
   \STATE{\mbox{Perform 2-dimensional convolutions on the minibatch}}
   \STATE{\mbox{Execute stochastic gradient descent on G}}
\ENDFOR

\end{algorithmic}
\end{algorithm*}


\subsection{Deep Convolutional GAN}
The DeepCancer architecture comprises a deep convolutional GAN inspired by the model in~\cite{Radford}. This is based on LeCun's model~\cite{Bottou}, which is most often employed in the supervised learning literature. As explained in Algorithm 1, we initially accept a 100-dimensional noise vector as the input to the generator. This is reshaped, batch-normalized and upsampled to a small spatial extent convolutional representation that goes through multiple convolutions with many activation maps. Finally, the high-level representation converts the input to a 224x183 dimensional map. On the other hand, this map acts as the input to the discriminator, which progressively convolves over the activation maps. These maps are flattened at the fully connected layer and passed through a sigmoid activation in order to carry out binary classification of samples.

For our generator model, batch normalization was employed for the model to learn features from samples passed in batches. Rectified Linear Units (ReLU) activations~\cite{Nair} were used. For our discriminator model, we employed LeakyReLU~\cite{Xu-Bing} activations and a dropout rate of 0.50.

A fundamental reason why DeepCancer is based on deep convolution networks is because of their excellent ability to learn features over different regions of the incoming input, while reducing the number of free parameters, which may invariably increase drastically in case of deep architectures. While our dataset doesn't explicitly demand deep architectures (networks with far more convolutional layers) like in 'ResNet'~\cite{Kaiming}, our adversarial model presents the scope for excellent results even in the case of larger datasets that will ultimately involve much more free parameters. 

Furthermore, convolution networks involve the neurons looking only at the convolving local receptive areas, and hence learning different features from different portions of the weight matrices, thereby strengthening the generalization ability of the model.

\section{Evaluation}
Two datasets have been utilized for testing DeepCancer. To introduce some terminology, `Alpha' refers to the learning rate, `Logistic\_C' refers to the regularizer of the Logistic Regression model, `svm\_C', refer to the cost regularizer of the SVM, `gamma' refers to the penalty constant of the SVM, `NoC' refers to the number of components, i.e., hidden units of the RBM. Also, in the GAN models, `AlphaD' and ‘AlphaG’ refer to the learning rates of the discriminator and the generator respectively.

\subsection{Datasets}

\subsubsection{GSE45584 Breast Cancer Dataset}
The dataset used for evaluating the model's ability to classify inflammatory breast cancer (IBC) samples is the accession ID GSE45584 dataset, which was created in~\cite{Woodward} and is currently part of the Gene Expression Omnibus data repository that houses curated microarray gene expression profiles for research purposes. The 45 samples (20 IBC, 20 non-IBC and 5 normal tissues) and their corresponding accession IDs are tabulated in Table~\ref{table:dataId}. For each of these samples, we have 41000 unique genes whose expression levels are positive floating-point real values that collectively indicate the activity, i.e., the expression of these several thousand genes in cellular function. Since genes contain instructions for messenger RNA (mRNAs), the extent of their contributions can be treated as the gene either being ‘on’ or ‘off’ while producing that mRNA. Furthermore, since numerous external factors decide whether or not a gene produces an mRNA, these expression profiles help in understanding the cell state, type, or environment among other attributes. Deep generative learning models understand these expressions, probabilistically firing neurons as per the expression level based on the aforementioned deep architecture that is capable of tackling high data dimensionality with minimal feature engineering, unlike traditional machine learning or hierarchical clustering approaches that depend on statistical tests to extract the most important features before performing classification.

A key aspect of the paper is the size of this dataset. Inflammatory breast cancer is rare~\cite{Dawood} and is characterized by the rapid onset of the erythema and swelling of the breast without any obvious breast mass. Therefore, owing to this characteristic of inflammation, diagnosis becomes extremely tedious, making retrieval of IBC samples difficult to obtain. Hence, the dataset compares gene expression levels of only twenty microdissected IBC samples with twenty microdissected non-IBC samples based on the marker status on the Human Epidermal Growth Factor Receptor (HER2/neu) amplification, and the Estrogen Receptor (ER)~\cite{Woodward}, apart from including five non-cancerous samples.

\subsubsection{Prostate Cancer Dataset}
The other dataset used in this paper is the prostate cancer dataset used in~\cite{Dinesh}, which involves fifty-two cancerous and fifty non-cancerous prostate specimens. In this case, for each sample, expressions of 12600 unique genes collectively indicated the gene activity. An important difference is that, unlike the GEO ID GSE45584 dataset, the expression levels were floating-point positive and negative real values. The central idea here is the same - profiling the expression levels indicate cellular activity and attributes like state. A similar procedure that was applied to the first dataset was also applied to the second dataset, with the evaluation metrics recorded being the same. Similar to the first dataset, the model segregated the samples into two classes – cancerous and non-cancerous prostate samples.

\begin{table*}[htb]
\renewcommand{\arraystretch}{1.5}
\caption{\textbf{Accession IDs of the 20 IBC, 20 non-IBC and 5 non-cancerous samples from the raw GSE45584 dataset}}
\label{table:dataId}
\centering 
\begin{tabular}{c|c|c|c|c|c|c}
\hline
 \textbf{Serial No.}&\multicolumn{2}{l|}{\textbf{Non-IBC}}&\multicolumn{2}{l|}{\textbf{IBC}}&\multicolumn{2}{l}{\textbf{Non-Cancerous}}\\
& \textbf{Sample} & \textbf{Geo\_accession ID}&  \textbf{Sample} & \textbf{Geo\_accession ID} & \textbf{Sample} & \textbf{Geo\_accession ID}\\
\hline
1&LGt008 & GSM1110043 & MB001 & GSM1110063 & LGp02NT & GSM1110083\\
2&LGt010 & GSM1110044 & MB002 & GSM1110064 & LGp03NT & GSM1110084\\
3&LGt024 & GSM1110045 & MB003 & GSM1110065 & LGp05NT & GSM1110085 \\
4&LGt025 & GSM1110046 & MB004 & GSM1110066 & LGp07NT & GSM1110086 \\
5&LGt032 & GSM1110047 & MB005 & GSM1110067 & LGp09NT & GSM1110087\\
6&LGt033 & GSM1110048 & MB006 & GSM1110068 & & \\
7&LGt038 & GSM1110049 & MB008 & GSM1110069 & & \\
8&LGt052 & GSM1110050 & MB009 & GSM1110070 & & \\
9&LGt058 & GSM1110051 & MB010 & GSM1110071 & & \\
10&LGt066 & GSM1110052 & MB014 & GSM1110072 & & \\
11&LGt067 & GSM1110053 &MB016 & GSM1110073 & & \\
12&LGt075 & GSM1110054 & MB018 & GSM1110074 & & \\
13&LGt082 & GSM1110055 & MB022 & GSM1110075 & & \\
14&LGt099 & GSM1110056 & MB025 &GSM1110076 & & \\
15&LGt107 & GSM1110057 & MB026 &GSM1110077 & & \\
16&LGt126 & GSM1110058 &MB028 & GSM1110078 & & \\
17&LGt131 & GSM1110059 &MB029 & GSM1110079 & & \\
18&LGt134 & GSM1110060 & MB030 & GSM1110080& & \\
19&LGt137 & GSM1110061 & MB031 & GSM1110081 & & \\
20&LGt141 & GSM1110062 & MB032 &GSM1110082 & & \\
\hline
\end{tabular}
\end{table*}

\subsection{Training}

The RBM-based models for the breast cancer dataset were tested on two different types of tasks:

\begin{itemize}
  \item Task 1: Classification of breast cancer tissues as either being non-inflammatory or inflammatory
  \item Task 2: Classification of tissues as either being cancerous and non-cancerous
\end{itemize}

\noindent whereas the prostate cancer classification involved only one task: classification of tissues as either being cancerous and non-cancerous.\\
\indent To accommodate for the first task, the five non-cancerous samples were removed during the preprocessing stages in order to ensure that only tissues diagnosed as cancerous were used by all the models. For the second task, the five normal breast tissues were augmented 8 times to ensure dataset balancing and prevent model training bias. These augmented samples formed the non-cancerous class, and were then concatenated with all the 20 non-IBC and 20 IBC samples that were grouped together to yield the cancerous class. Therefore, for the first task, the size of the input matrix after preprocessing was 40 X 40992, accommodating 20 samples each of the IBC and non-IBC categories, whereas for the second task, this size after dataset augmentation was 80 X 40992. The original prostate cancer dataset has two separate files as training and test data, with 25 prostate samples and only 9 non-prostate samples in the test data, both these files were combined to obtain an input matrix of size 136 X 12600. Prior to training, samples were shuffled and then split into training and testing sets to avoid any model bias or overfitting.

Similarly, for DeepCancer, we recognized three different tasks:

\begin{itemize}
  \item Task 1: Classification of breast cancer tissues as either being cancerous and non-cancerous.
  \item Task 2: Classification of breast cancer tissues as either being inflammatory or non-inflammatory.
  \item Task 3: Classification of prostate cancer tissues as either being cancerous and non-cancerous.
\end{itemize}

Similar to the RBM-based models, to carry out these GAN-based tasks, three tasks were defined as tabulated in Table \ref{table:task1}.

\begin{table*}[htb]
\renewcommand{\arraystretch}{1.5}
\caption{\textbf{Three different tasks with sub-tasks defined for evaluation of the convolutional GAN model}}
\label{table:task1}
\centering 
\begin{tabular}{l c c c c}
 \hline
 \textbf{Task} & \textbf{Part-I} && \textbf{Part-II}    \\
 \hline
 & \textbf{a} & \textbf{b} & \textbf{a} & \textbf{b}    \\
 \hline

\textbf{Task1} & Tr:Cancerous, Ts:Cancerous & Tr:Cancerous,Ts:Non-Cancerous &Tr:Non-Cancerous,Ts:Non-Cancerous  & Tr:Non-Cancerous, Ts:Cancerous\\
\textbf{Task2} & Tr:IBC,Ts:IBC & Tr:IBC, Ts:Non-IBC & Tr:Non-IBC,Ts:Non-IBC & Tr:Non-IBC,Ts:IBC \\
\textbf{Task3} & Tr:Prostate,Ts:Prostate & Tr:Prostate, Ts:Non-Prostate & Tr:Non-Prostate,Ts:Non-Prostate & Tr:Non-Prostate,Ts:Prostate \\
\hline
\end{tabular}
\end{table*}

\noindent Size of the input matrices for DeepCancer model was similar to that used in the ID GSE45584 dataset. Different training and test split proportions were tested, as mentioned in the subsequent sections below. \\
\indent A key aspect of the paper is the size of the ID GSE45584 dataset. Inflammatory breast cancer is quite a rare form of cancer~\cite{Dawood} and extremely tedious to diagnose, making retrieval of inflammatory breast cancer samples difficult to obtain. Therefore, while the paper focusses on a smaller dataset, the authors believe that the model should perform equally well for bigger datasets.

Table \ref{table:ganarch} represent the architecture design of our GAN model; the GSE45584 and the prostate cancer dataset hyperparameters are separated by `/'. 

\begin{table}[htb]
\renewcommand{\arraystretch}{1.0}
\caption{\textbf{Convolutional GAN Architecture Parameters}}
\label{table:ganarch}
\centering
\begin{tabular}{l c c c c}
 \hline
 \textbf{Layer} & \textbf{Kernel length/} & \textbf{Number of Kernel/}\\
  & \textbf{Pool Size*} & \textbf{Number of Neurons*} \\
 \hline
 \textbf{\underline{Generator Module}} &  &   \\
 \textbf{Dense1} & - & 40992*/12600* &  \\
\textbf{UpSampling1} & (32,3)*/(10,10)* &   \\
\textbf{Convolution2} & 3/3 & 48/50 &  \\
\textbf{Convolution3} & 3/3 & 32/25 &  \\
\textbf{Convolution4} & 1/1 & 1/1 &  \\
\\
\textbf{\underline{Discriminator Module}} &  &    \\
\textbf{Convolution5} & 3/3 & 32/32 & \\
\textbf{Convolution6} & 3/3 & 64/64 & \\
\textbf{Dense2} & - & 64*/64* &  \\
\textbf{Dense3} & - & 2*/2* &  \\
\hline
\end{tabular}
\end{table}

\subsection{Evaluation Metrics}
Since the nature of the task is human-centric, i.e., to classify breast cancer tissues based on gene signatures, the most appropriate metric that preserves this nature is the precision-recall metric. Therefore, conforming to statistical postulates, a high precision and high recall imply that the model is able to correctly classify all the detected samples. A high precision with 
low recall implies that the model detects fewer samples but classifies almost all of them correctly. Whereas, lower precision with higher recall implies that the samples are being misclassified. This is particularly important to understand since, pertaining to the nature of the task, a non-inflammatory breast cancer sample classified as an inflammatory sample is medically inaccurate but ultimately tolerable as compared to the highly undesirable scenario of inflammatory samples classified as non-inflammatory, given that inflammation poses far more serious medical consequences.

\begin{table}[htb]
\renewcommand{\arraystretch}{1.5}
\caption{\textbf{Learning rate of the RBM-SVM model examined on the GSE45584 dataset with the NoC = 200, Split=0.5, svm\_C=1, Epochs=50, gamma=0.06}}
\label{table:rbmsvm}
\centering
\begin{tabular}{c c c c}
 \hline
 \textbf{Alpha} & \textbf{Precision} & \textbf{Recall} & \textbf{F1-Score}\\
 \hline
\textbf{0.0006} & 79 & 65 & 60\\
\textbf{0.0009} & 83 & 75 & 74 \\
\textbf{0.001} & 71 & 70 & 71 \\
\textbf{0.005} & 86 & 80 & 79 \\
\textbf{0.01} & 78 & 60 & 52 \\
\hline
\end{tabular}
\end{table}

\begin{figure}[!ht]  
\centering
\includegraphics[scale=0.65]{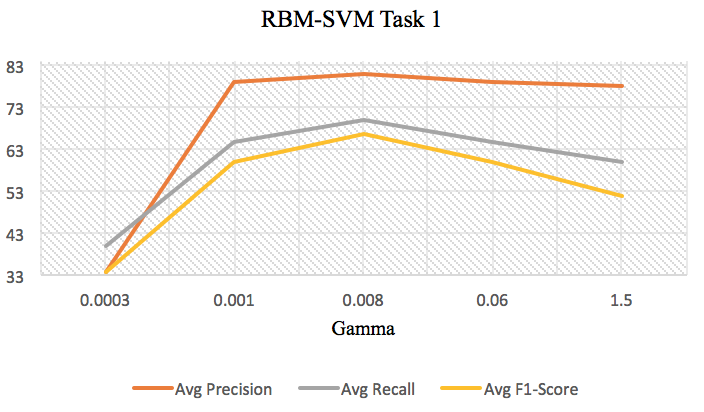}  
\caption{Variations in evaluation scores due to gamma on the GSE45584 dataset for the RBM-SVM model}
\label{fig:rbmsvm}
\end{figure}

\begin{figure}[!ht]  
\centering
\includegraphics[scale=0.65]{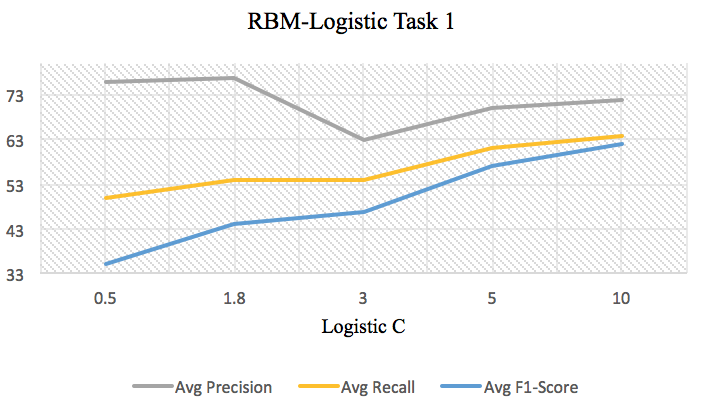}  
\caption{Variations in evaluation scores due to regularizer C on the GSE45584 dataset for the RBM-Logistic model}
\label{fig:rbmlogistic}
\end{figure}

\begin{table}[htb]
\renewcommand{\arraystretch}{1.5}
\caption{\textbf{gamma of the RBM-SVM model examined on the GSE45584 dataset with NoC = 200, Split=0.5, svm\_C=1, Epochs=50, alpha=0.0006.}}
\label{table:rbmsvm_model}
\centering
\begin{tabular}{c c c c}
 \hline
 \textbf{Gamma} & \textbf{Precision} & \textbf{Recall} & \textbf{F1-Score}\\
 \hline
\textbf{0.0003} & 34 & 40 & 34\\
\textbf{0.001} & 79 & 65 & 60 \\
\textbf{0.008} & 81 & 70 & 67 \\
\textbf{0.06} & 79 & 65 & 60 \\
\textbf{1.5} & 78 & 60 & 52 \\
\hline
\end{tabular}
\end{table}

\subsection{Model Performance}
In order to appropriately test the efficacy of the system in the presence of different hyperparameters, only one of them was varied while all the others were fixed. These fixed hyperparameters were recorded after multiple rounds of model selection and the set of values that yielded the most optimal results have been used. 

\begin{figure*}[!ht]  
\centering
\includegraphics[scale=0.55]{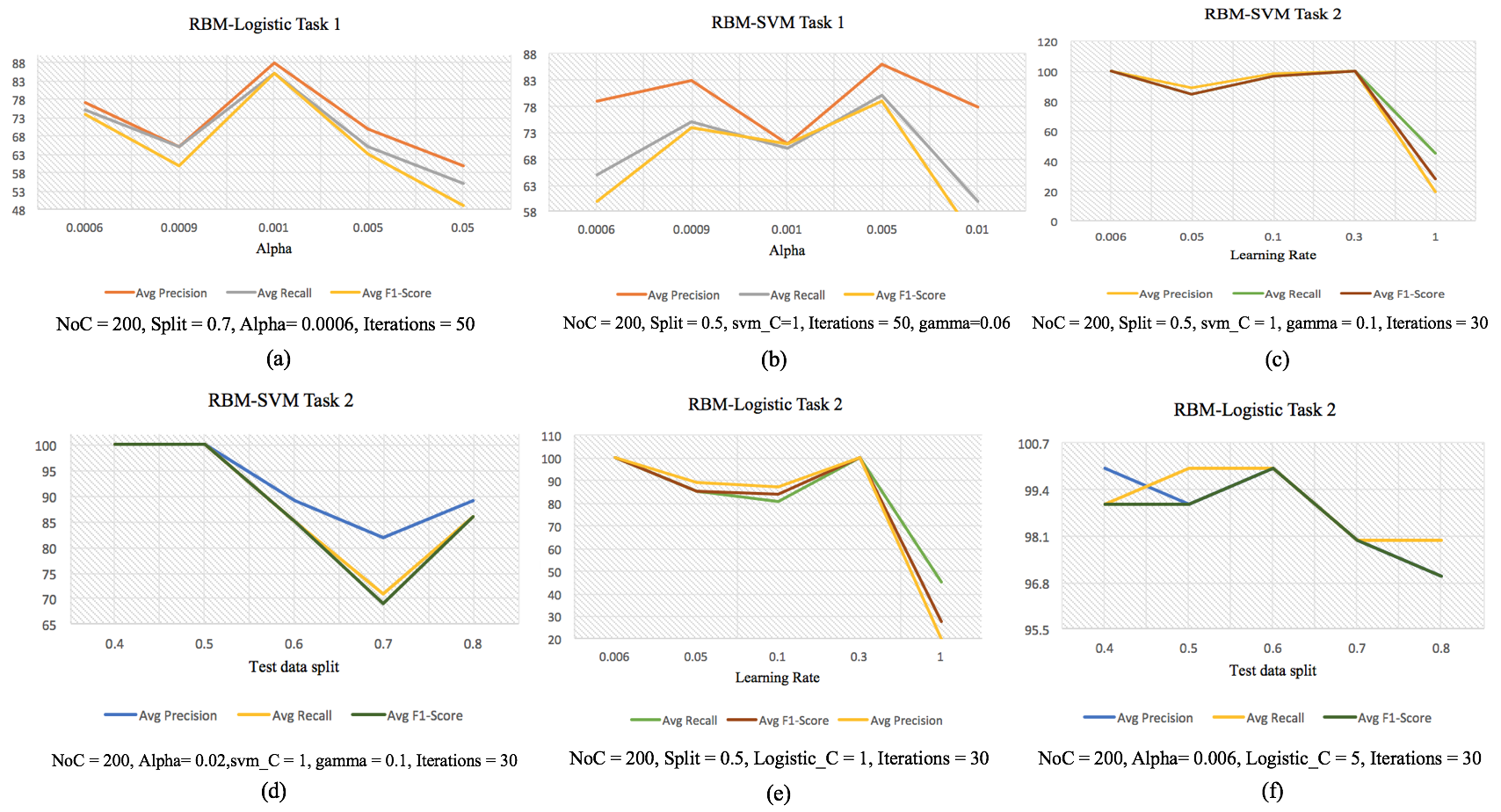}  
\caption{Effects of different configurations of the model on the Average Precision, Recall and F1-scores}
\label{fig:model_stat}
\end{figure*}

Based on the aforementioned implementation, training and evaluation procedures, the baseline RBM-Logistic Regression model on the GSE45584 dataset, was able to detect 20 samples: 12 non-IBC and 8 IBC. Its performance, as shown in Fig. \ref{fig:model_stat}(a), was significantly favorable for differentiating non-IBC samples from IBC samples with 88\% (approximately 11 out of 12 samples) model average precision score, with an associated average recall score of 85\%, thereby indicating that the model learns to classify an IBC tissue accurately.

Also, the learning rate hyperparameter of the RBM is a key component that governs how effectively convergence occurs in the model. The best precision score was observed for a learning rate of 0.001 with the precision-recall curve eventually dropping sharply below 0.001, further suggesting the condition of optimality in feature learning. Another key hyperparameter is the model pipeline is logistic regularizer, C, that is traditionally used to avoid 'underfitting' (model bias) by learning many features. Basically, some features may be assigned higher weights than others, thus biasing the model and reducing precision. To avoid this, features with extremely high weights are rejected while ensuring effective learning and smooth convergence to local minima. Therefore, balancing this trade-off between effective feature learning and higher precision is crucial to our task, since lesser learning may impact the overall performance of the system. As Fig. \ref{fig:rbmlogistic} shows, the most optimal C was experimentally found to be 1.8 for an average precision of 77\% (approximately 10 out 13 samples). Interestingly, the corresponding recall score of 54\% once again validates the efficacy of our model in terms of accurate classification to a good extent.

The RBM-SVM model (Tables \ref{table:rbmsvm} and \ref{table:rbmsvm_model}) detected 10 IBC and 10 non-IBC samples and performed equally well, reporting 86\% (approximately 9 out 10 samples) as the highest average precision score with the recall score being 80\% implying accurate classification of almost all the detected IBC samples. The learning rate of the RBM was most optimal for 0.005 with the steepest drop in precision thereafter. Likewise, the penalty term, $\gamma$ led to significant increase in model precision after much experimentation. While, $\gamma=0.0003$ reported a precision of just 34\%, fine-tuning it to the range 0.001--0.06 presented a much higher precision range. Also, recall scores recorded for variations in $\gamma$ led to higher recall. Overall, the model was able to pick up highly relevant features and classify almost all the detected samples accurately.

On the second task, the RBM-Logistic model yielded an average precision of 89\% (approximately 17 out of 18 IBC) samples for $\alpha=0.05$ (Fig. \ref{fig:model_stat}(e)), while reporting an equally high recall of 85\%. Furthermore, tweaking to $\alpha=0.03$ resulted in all detected IBC samples being accurately classified. Other parameters such as train-test data split was also experimentally tested to record observations. 40--60\% of the dataset assigned as the test set proved to be most effective for learning.

\begin{figure*}[!ht]  
\centering
\includegraphics[scale=0.55]{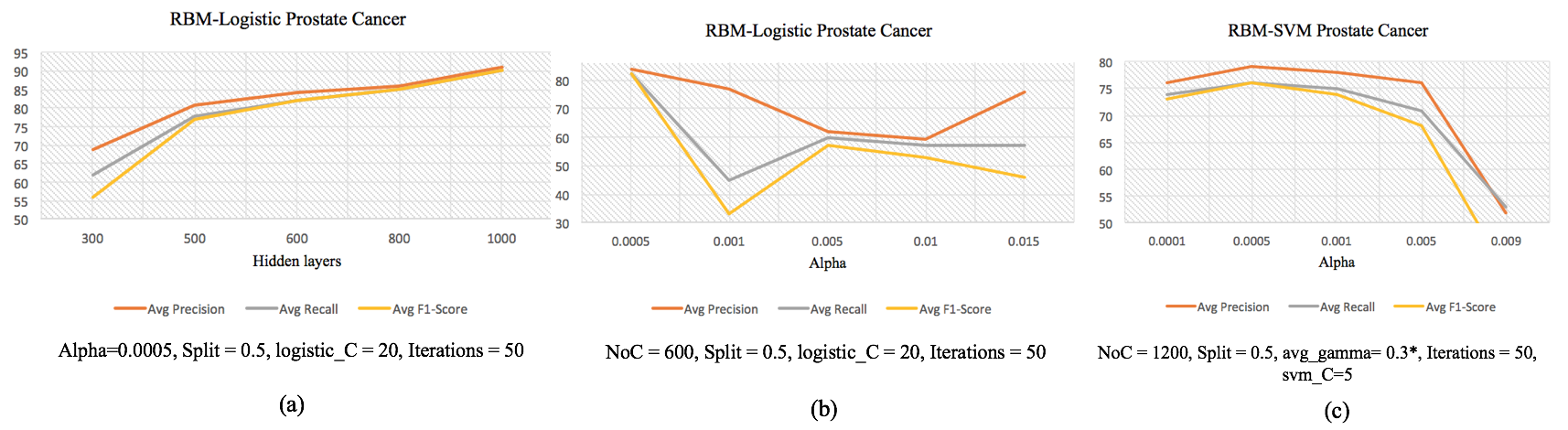}  
\caption{Effects of different configurations of the model on the Average Precision, Recall and F1-scores}
\label{fig:model_config}
\end{figure*}

Similarly, for the prostate cancer dataset, the RBM-Logistic Regression model performed very well with increased number of hidden neurons performing exceedingly well in differentiating the cancerous samples from the non-cancerous samples (Fig. \ref{fig:model_config}(a)). A direct relationship between the number of hidden units and model precision is discernible, as the average precision increased steadily from 69\% using 300 neurons to 91\% using 1000 neurons. Compared to the SVM classifier in task 1, the model learnt to classify far better using $\alpha=0.0005$. Interestingly, the average precision improves after $\alpha=0.01$ but this was observed keeping NoC = 600. This may not be the case if the NoC were higher.

The performance of DeepCancer was far better than the baseline model, with the generator learning to accurately represent the features of the GSE45584 dataset and imitating the characteristics well enough to be discriminated favorably by the discriminator, i.e., the discriminator accepts these samples after several epochs of discrimination between real and fake samples, thereby improving in its ability to classify samples into their appropriate categories. 

\begin{figure*}[!ht]  
\centering
\includegraphics[scale=0.55]{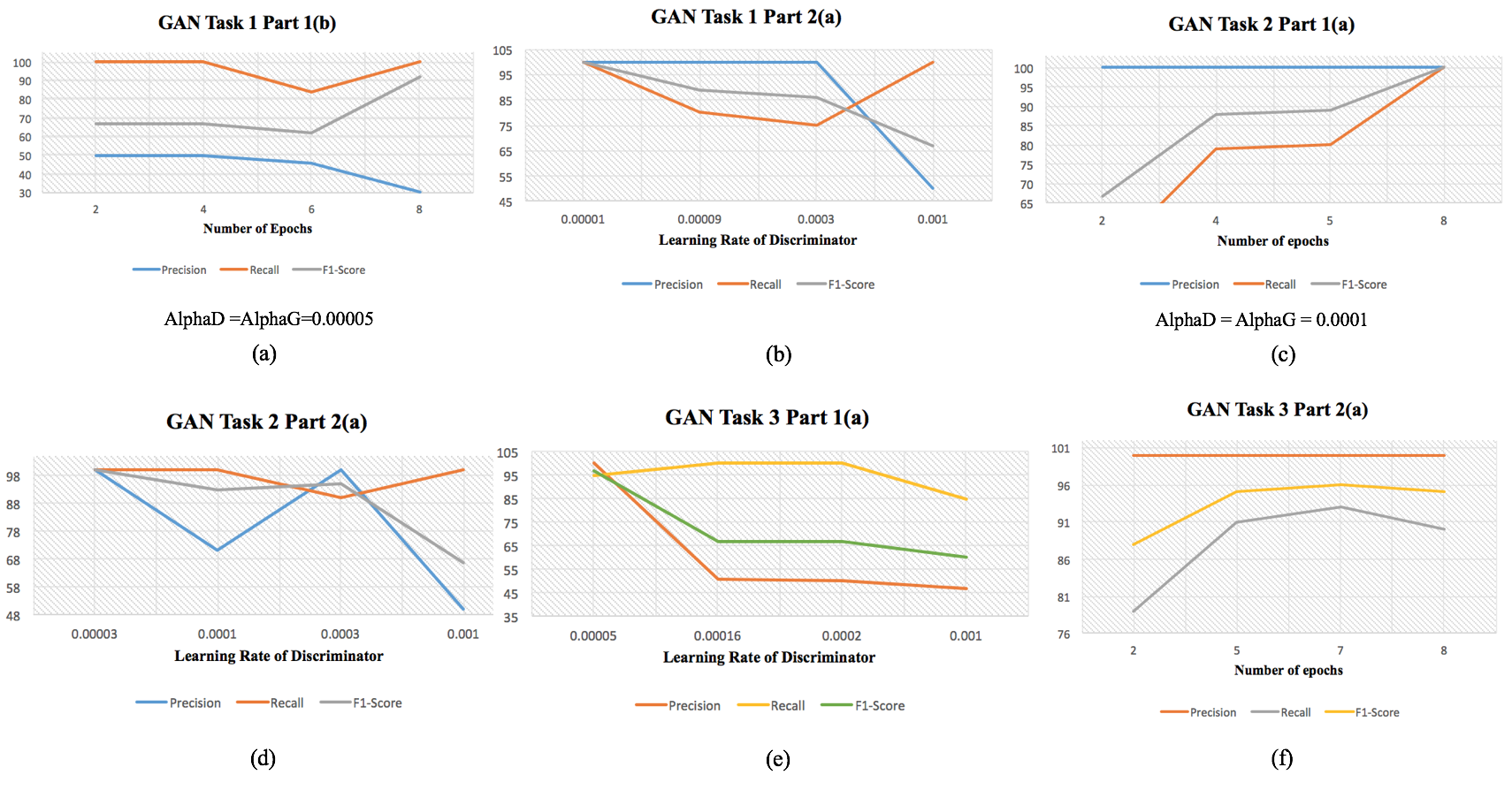}  
\caption{Evaluation of the convolutional GAN based on varying GAN-sensitive hyperparameters}
\label{fig:gan_model2}
\end{figure*}

A general practice in computationally-heavy models is to pre-train the networks in order to reduce the overall computation time. However, due to a dataset relatively smaller than popular datasets such as the MNIST~\cite{Bottou}, pre-training was observed to be an essential but avoidable step. We exhaustively experimented with the learning rate of the discriminator and examined the effect of epochs on precision-recall and reported this performance (Fig. \ref{fig:gan_model2}). On Task 1, the learning rate of the discriminator for the range $\alpha=0.00001$ to $\alpha=0.001$  (Fig. \ref{fig:gan_model2}(b)). Very clearly, as the learning rate was increased, the precision-recall pair of the model starts to decrease in 10 percentage-points, and finally misclassifies half of the samples for the highest value of $\alpha=0.001$. This is in line with machine learning theory, since greater the learning rate of the model, the more difficult it is for the model to achieve convergence while updating the gradient descent parameters. Additionally, F1-Scores are a good indicative of the general overall performance of the model, since it considers the harmonic mean of the precision and recall scores. On Task 1, our model reports an overall average F1-Score of 84\%, indicating a favorable performance against the objective of two-class classification. 

Similarly, for Task 2, Fig. \ref{fig:gan_model2}(d) shows that as the learning rate was increased, the performance of the model deteriorates with the model going from classifying accurately at $\alpha=0.00003$ to misclassifying half the samples at $\alpha=0.001$. An interesting phenomenon that we repeatedly observed was a sudden spike in the precision for $\alpha=0.0003$. This, the authors believe, was probably because of the gradient getting stuck in local minima during gradient descent, a persistent problem in machine learning while descending quickly. For these observations, the F1-scores falls drastically from 100\% to about 67\% as the algorithm undergoes gradient descent. Also, the precision-recall scores obtained for Task 2 part 2(b) were expected to be far apart, considering the nature of the task: to classify non-IBC from IBC samples after training on non-IBC samples and testing on IBC samples (Table \ref{table:gan_model1}).

\begin{table}[htb]
\renewcommand{\arraystretch}{1.5}
\caption{\textbf{Effects of varying the learning rate of the discriminator in Task 2 part 2(b)}}
\label{table:gan_model1}
\centering
\begin{tabular}{c c c c}
 \hline
 \textbf{AlphaD} & \textbf{Precision} & \textbf{Recall} & \textbf{F1-Score}\\
 \hline
\textbf{0.00003} & 55 & 100 & 70\\
\textbf{0.0001} & 50 & 100 & 67 \\
\textbf{0.0003} & 30 & 95 & 45 \\
\textbf{0.001} & 25 & 100 & 40 \\
\hline
\end{tabular}
\end{table}

\begin{table}[htb]
\renewcommand{\arraystretch}{1.5}
\caption{\textbf{Effects of varying the learning rate of the discriminator in Task 3 part 2(b)}}
\label{table:gan_model2}
\centering
\begin{tabular}{c c c c}
 \hline
 \textbf{Epochs} & \textbf{Precision} & \textbf{Recall} & \textbf{F1-Score}\\
 \hline
\textbf{2} & 50 & 100 & 67\\
\textbf{4} & 45 & 100 & 62 \\
\textbf{6} & 30 & 95 & 45 \\
\textbf{8} & 25 & 100 & 40 \\
\hline
\end{tabular}
\end{table}

GAN models are inherently difficult to train~\cite{Radford} and require number of epochs for generation-based tasks. To test this, the number of training and testing epochs was experimentally modified to study its impact. On Task 3 part 1(a), 2 epochs led to a precision of just 79\% (Fig. \ref{fig:gan_model2}(f)), and a steady increase in accuracy was observed as the epochs were increased to 8. F1-Scores of these reading also indicate that higher epochs generally led to better results. Similarly, on Task 3, while 2 epochs reported a precision of 100\%, the recall was much lower (79\%), therefore motivating us to experiment further. High accuracy on this task is no surprise since it measures model performance after training and testing on the non-prostate tissues.





\section{Conclusion}
We presented a deep generative learning model DeepCancer for detection and classification of inflammatory breast cancer and prostate cancer samples. The features are learned through an adversarial feature learning process and then sent as input to a conventional classifier specific to the objective of interest. After modifications through specified hyperparameters, the model performs quite comparatively well on the task tested on two different datasets.

The proposed model utilized cDNA microarray gene expressions to gauge its efficacy. Based on deep generative learning, the tuned discriminator and generator models, D and G respectively, learned to differentiate between the gene signatures without any intermediate manual feature handpicking, indicating that much bigger datasets can be experimented on the proposed model more seamlessly. The DeepCloud model will be a vital aid to the medical imaging community and, ultimately, reduce inflammatory breast cancer and prostate cancer mortality.

\bibliographystyle{IEEEtran} 

\bibliography{references}

\end{document}